\pdfoutput=1
\documentclass{bmvc2k}

\usepackage{times}
\usepackage{epsfig}
\usepackage{graphicx}
\usepackage{amsmath}
\usepackage{amssymb}
\usepackage{booktabs}
\usepackage{multirow}
\usepackage{xcolor}
\usepackage{mathtools}
\usepackage{dsfont}
\usepackage{floatrow}
\newfloatcommand{capbtabbox}{table}[][\FBwidth]
\DeclareMathOperator*{\argmax}{argmax}

\DeclareRobustCommand\onedot{\futurelet\@let@token\@onedot}
\def\@onedot{\ifx\@let@token.\else.\null\fi\xspace}

\title{Semi-Supervised Learning with Taxonomic Labels}

\addauthor{Jong-Chyi Su}{jcsu@cs.umass.edu}{1}
\addauthor{Subhransu Maji}{smaji@cs.umass.edu}{1}

\addinstitution{
 University of Massachusetts Amherst\\
 Amherst, MA, USA
}

\runninghead{Su, Maji}{Semi-Supervised Learning with Taxonomic Labels}

\def\eg{\emph{e.g}\bmvaOneDot}
\def\ie{\emph{i.e}\bmvaOneDot}

\def\etal{\emph{et al}\bmvaOneDot}

\begin{document}

\maketitle

\begin{abstract}
We propose techniques to incorporate coarse taxonomic labels to train image classifiers in fine-grained domains. Such labels can often be obtained with a smaller effort for fine-grained domains such as the natural world where categories are organized according to a biological taxonomy. On the Semi-iNat dataset consisting of 810 species across three Kingdoms, incorporating Phylum labels improves the Species level classification accuracy by 6\% in a transfer learning setting using ImageNet pre-trained models. Incorporating the hierarchical label structure with a state-of-the-art semi-supervised learning algorithm called FixMatch improves the performance further by 1.3\%. The relative gains are larger when detailed labels such as Class or Order are provided, or when models are trained from scratch. However, we find that most methods are not robust to the presence of out-of-domain data from novel classes. We propose a technique to select relevant data from a large collection of unlabeled images guided by the hierarchy which improves the robustness. Overall, our experiments show that semi-supervised learning with coarse taxonomic labels are practical for training classifiers in fine-grained domains. 
\end{abstract}

\section{Introduction}\label{sec:intro}
Large labeled datasets have been the key to the success of deep networks for many tasks. However, labeling requires expertise and can be time-consuming, especially for fine-grained recognition tasks such as identifying the species of birds~\cite{WahCUB_200_2011} or variants of aircrafts~\cite{maji2013fine}.
In this work, we investigate if coarsely labeled datasets can be used to improve performance of a target fine-grained recognition task. 
For example, in natural domains one can often obtain a large dataset with the same coarse labels as the target task through community driven platforms such as iNaturalist~\cite{van2018inaturalist}.
Effectively incorporating them in a semi-supervised learning framework to improve performance could be a compelling alternative to existing few-shot learning approaches, which have been less effective in fine-grained domains~\cite{su2021realistic,cole2021when}.

We present an analysis on the Semi-iNat dataset~\cite{su2021semi_iNat} that consists of images from 810 species spanning three Kingdoms and eight Phyla (Figure~\ref{fig:main} and Table~\ref{tab:taxa}). The dataset contains: (i) a small set of images labeled at the species level (in-class), (ii) a large set (9$\times$) of coarsely-labeled images from the same species (in-class), and (iii) an even larger set (32$\times$) of coarsely-labeled images from novel species within the same taxonomy (out-of-class). At test time, the species classification accuracy is measured on novel images of the set of species within the labeled set (in-class).
The dataset is fine-grained and long-tailed, posing challenges to existing approaches for semi-supervised learning.

\definecolor{myblue}{RGB}{48,118,181}
\definecolor{mygreen}{RGB}{84,173,50}
\definecolor{myorange}{RGB}{227,120,46}
\definecolor{mypurple}{RGB}{136,71,246}
\begin{figure}[t!]
\centering
\includegraphics[width=0.99\linewidth]{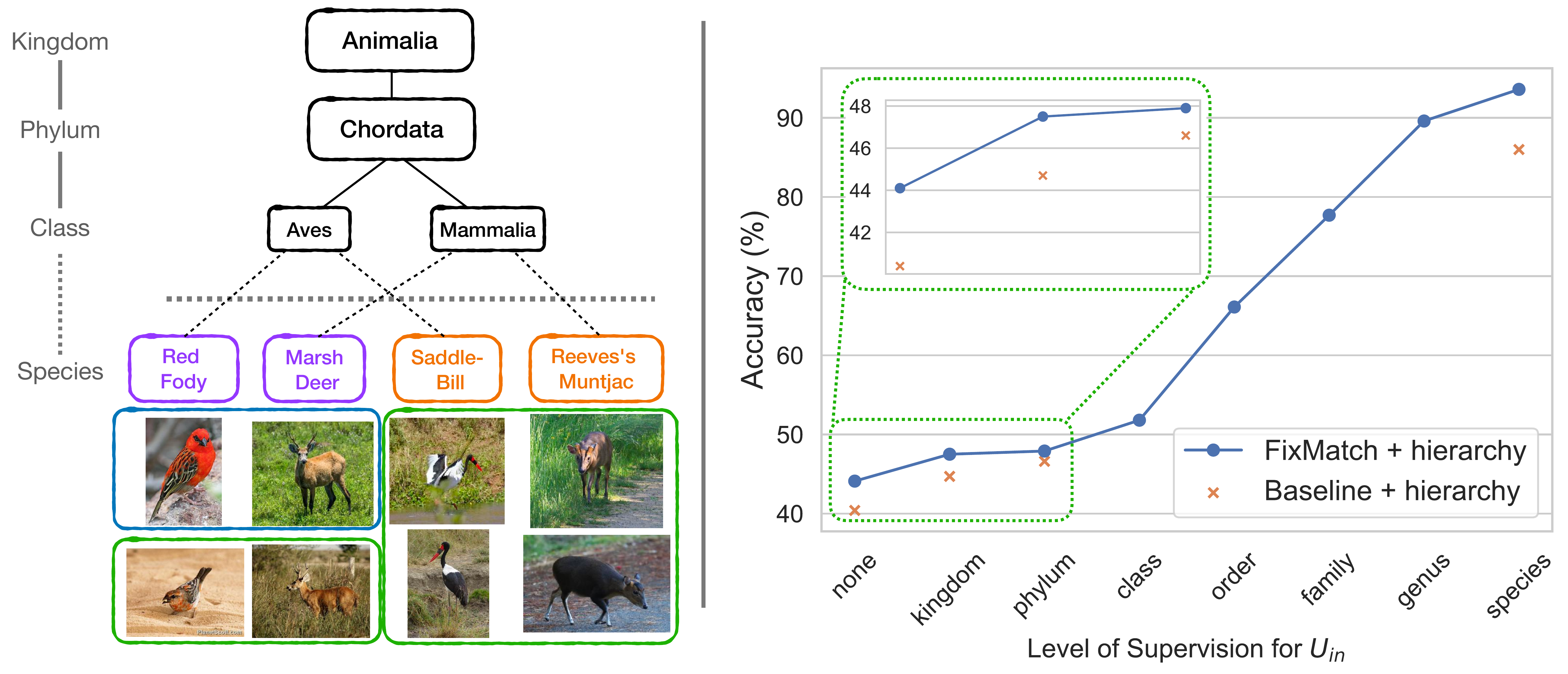}
\caption{\textbf{Taxonomic labels improve semi-supervised learning.} \textbf{Left:} The Semi-iNat dataset~\cite{semi_inat_challenge} contains \textcolor{myblue}{labeled data} from \textcolor{mypurple}{in-class} Species and \textcolor{mygreen}{coarsely labeled data} from both \textcolor{mypurple}{in-class} and \textcolor{myorange}{out-of-class} Species. The coarse labels such as Kingdom and Phylum, can be derived from the taxonomy given the Species, but are easier to annotate. \textbf{Right:} We observe that incorporating coarse labels improves the Baseline and FixMatch across different levels of supervision. FixMatch provides additional gains over the supervised Baseline which is based on fine-tuning an ImageNet pre-trained ResNet-50 using a hierarchy-aware loss (\S~\ref{sec:method}).}
\label{fig:main}
\end{figure}

For a consistent evaluation, we present results using a ResNet-50 network trained from scratch or pre-trained on ImageNet~\cite{deng2009imagenet} using various semi-supervised and self-supervised learning approaches.
For the supervised Baseline with ImageNet pre-training, incorporating a hierarchical loss at the Phylum level consisting of eight categories improves the Top-1 accuracy from 40.4\% to 46.6\% (Figure~\ref{fig:main} and Table~\ref{tab:benchmark_hie_in}).
This beats the gains using semi-supervised learning with FixMatch~\cite{sohn2020fixmatch} alone, which obtains 44.1\%. 
However, the gains are complementary and combining the two improves the performance to 47.9\%. 
Coarse taxonomic labels are also useful when models are trained from scratch, improving the performance of the best method from 32.0\% to 34.5\% (Table~\ref{tab:benchmark_hie_in}).
Figure~\ref{fig:main} quantifies the gains obtained with the Baseline and FixMatch using supervision from the Kingdom level (3 categories) to the Species level (810 categories, full supervision). Kingdom and Phylum supervision provides consistent gains for both methods, and improves over the semi-supervised learning without the coarse labels. 
The benchmark is far from saturated as indicated by the performance of the fully-supervised Baseline at roughly 86\%.

A common assumption in semi-supervised learning is that the weakly-labeled data belongs to the same set of classes as the target task. This is hard to guarantee in practice as coarsely labeled datasets collected in-the-wild may contain novel classes.
We find that the presence of this out-of-domain data leads to a drop in performance for nearly all approaches.
For example, the performance of FixMatch drops from 47.9\% to 41.1\% (Table~\ref{tab:benchmark_hie_out}) when the larger set of images are included for semi-supervised learning. 
This is problematic because labeling if an image contains novel classes is significantly more challenging than obtaining a large pool of images with the same coarse labels.
We find that prior work based on importance weighting based on a domain classifier (\eg,~\cite{Su2020When}) is less effective, but using the hierarchy to exclude the novel categories leads to a small improvement in some cases.

In summary, we show that coarse labels improve supervised and semi-supervised learning in fine-grained natural taxonomies.
In particular, we find that:
(i) coarse labels can be incorporated in several state-of-the-art methods to boost their performance;
(ii) The improvements are greater with more fine-grained labels;
(iii) The presence of novel classes hurts performance, but this can be somewhat mitigated by techniques for detecting domain shifts. However, the marginal gains obtained in this setting highlights the difficulty of detecting novel classes in fine-grained domains. The code is available at \url{https://github.com/cvl-umass/ssl-evaluation}.

\section{Related Works}\label{sec:related}
\paragraph{Semi-supervised learning.} Semi-supervised learning aims to use weakly labeled data to improve the model generalization.
Self-training approaches proposed in some early work~\cite{mclachlan1975iterative,scudder1965probability} use the model's own prediction to generate labels.
Their modern incarnations include pseudo-labeling~\cite{lee2013pseudo} which uses confident predictions as target labels for unlabeled data. 
Such labels can also be added gradually as a form of curriculum learning ~\cite{bengio2009curriculum,cascante2020curriculum} to reduce model drift.
UPS~\cite{rizve2021defense} generalizes this idea and incorporates low-probability predictions as negative pseudo-labels for multi-label prediction tasks. 
Other methods use a combination of self-supervised and semi-supervised learning techniques~\cite{zhai2019s,gidaris2019boosting,Su2020When}, which is sometimes followed by an additional step where the model's predictions are used to train a ``student model'' using distillation~\cite{xie2020self,yalniz2019billion,zoph2020rethinking,chen2020big}. 
Consistency-based approaches enforce the similarity of predictions between two augmentations of the same data, or different models in an ensemble, as a form of regularization~\cite{bachman2014learning,rasmus2015semi,laine2016temporal,sajjadi2016regularization,miyato2018virtual}. 
The role of augmentations have been explored in detail in techniques such as  MixMatch~\cite{berthelot2019mixmatch}, ReMixMatch~\cite{berthelot2019remixmatch}, FixMatch~\cite{sohn2020fixmatch}, and UDA~\cite{xie2019unsupervised}, which combine geometric and photometric image augmentations with other techniques such as  MixUp~\cite{zhang2017mixup}.

\paragraph{Learning with hierarchical labels.}
The hierarchical structure of the label space can be used to improve classification performance in many ways~\cite{ristin2015categories,taherkhani2019weakly,guo2018cnn}.
A common approach is to frame the problem as a structured prediction task and use a graphical model that incorporates the label structure with standard Bayesian machinery.
For example, Deng~\etal~\cite{deng2014large} exploit the inclusion and exclusion relations  among the labels to improve classification. 
YOLOv2~\cite{redmon2017yolo9000} learns object detectors across a large set of categories by predicting labels on the taxonomy in a top-down manner by learning a conditional distribution of the leaves given the parent category. 
This approach parameterizes the learnable weights along the edges of the tree. We instead parameterize the weights along the leaves of the tree.
Other works predict fine-grained labels by designing models that predict the labels~\cite{yan2015hd,zhu2017b} or
learn features~\cite{wang2015multiple} across different levels in the hierarchy in a multi-task framework.
The hierarchical label space can also be utilized to detect novel classes. 
For example, given coarse labels, Hsieh~\etal~\cite{hsieh2019pseudo} learn to assign fine-grained pseudo-labels by meta-learning, assuming that the classifier can achieve the best performance when the missing labels are correctly recovered.
Another application is zero-shot learning where attributes of the novel classes are provided, but there are no training images in novel classes~\cite{samplawski2020zero,lee2018hierarchical}. 
Recently, coarse labels have been incorporated in contrastive learning to improve image retrieval~\cite{touvron2020grafit} and few-shot learning~\cite{bukchin2020fine,Phoo2021Coarsely}.
Unlike prior work, we investigate if hierarchical labels can be used for improving of semi-supervised learning, for example by constraining the label space of approaches such as FixMatch or Pseudo-Labeling.

\section{Method}\label{sec:method}
\paragraph{Notation and problem setting.}
We focus on a structured prediction task where the label space $y \in {\cal Y}$ has a hierarchical structure.
It corresponds to a tree-structured biological taxonomy with 7 levels corresponding to the \emph{Kingdom, Phylum, Class, Order, Family, Genus,} and \emph{Species}. 
Denote $y^{l}$ as the label of an instance at the level $l$. 
Thus, $y^{1}$ is the label at the Kingdom level, $y^{2}$ for the label in the Phylum level, and the leaf nodes in the tree correspond to Species-level labels denoted by $y^{7}$.
Similarly, denote the sets of label space at a level $l$ as $\mathcal{C}^{l}$. 
For example, the label space in the species level is $y^{7} \in \mathcal{C}^{7}$, and in the Phylum level is $y^{2} \in \mathcal{C}^{2}$. 
Given the label in a level, we can infer the labels in all the upper levels using the tree structure, \eg, we can infer the Kingdom label given the Phylum label.
The Semi-iNat~\cite{su2021semi_iNat} dataset provides Species-level labels for a subset of images, but coarse labels (\eg, Kingdom and Phylum) for a larger set of images (Table~\ref{tab:taxa}). Performance is measured as the accuracy at the Species level on novel images.

\paragraph{Model parameterization.} 
We train a model to predict the labels at the finest level, \ie, $y^7 = f(x)$, and marginalize over the tree structure to obtain probabilities at any level. There are several alternatives to this schema. For example, we could train separate heads for each level instead of just the leaves. However, this performed worse, obtaining 42.1\% accuracy on a supervised model trained from ImageNet, while our proposed method achieves 46.6\% accuracy. We could also consider the parameterization proposed in YOLOv2~\cite{redmon2017yolo9000} which models the conditional distribution of the leaves given the parent for each internal node in the tree. Both schemes require many more parameters. For example, there are 2041 edges in the taxonomy (summing over Table~\ref{tab:taxa} right); thus, the edge-parameterization of YOLOv2 would require 2041 weights (\#edges) compared to 810 weights (\#leaves). While the edge parameterization can handle arbitrary graphs, it offers no obvious advantage for tree-structured models. Having fewer weights may be preferable in the few-shot setting. 
\subsection{Hierarchical supervised loss}
We consider the supervised cross-entropy loss in each level of the hierarchy. 
For labeled data $(x_i,y_i^{7}) \in \mathcal{L}$, the model $f$ first predicts the label space of the species $\mathcal{C}^{7}$ with the probabilities $p_i^{7} = f(x_i)$. 
For coarse labeled data, say we have the label at the Phylum level $(u_j,y_j^{2}) \in \mathcal{U}$, we first use the same model to predict the labels in the species level $q_j^{7} = f(u_j)$. 
We then apply a cross-entropy loss over the model's prediction at the Phylum level obtained by summing the probabilities of all the leaf nodes under each Phylum.
The marginalization can be done by $q_j^{2} = q_j^{7} \cdot W_{7}^{2}$, where the predefined matrix $W_{7}^{2}$ represents the edges between the Species and Phylum level (the elements are 1 for the edges and 0 otherwise). 

During training, we sample $m$ labeled data from $\mathcal{L}$ and $n$ coarsely labeled data from $\mathcal{U}$ in each batch for stochastic gradient descent.
For labeled data, we only add supervised loss on the lowest level, which is the Species level.
The complete hierarchical supervised loss is:
\begin{equation}\label{eq:hierarchy}
\mathcal{L}_\text{hie}^{7,2} = \sum_{i=1}^{m} H(y_i^{7}, p_i^{7}) + \sum_{j=1}^{n} H(y_j^{2}, q_j^{2}) ,
\end{equation}
where $H$ is the cross-entropy function $H(u,v) = -\sum_w{u(w)\log v(w)}$. 
The first term is the loss for labeled data on the Species level, and the second term is the loss for coarsely labeled data on the Phylum level. 
The superscript on the loss $\mathcal{L}_\text{hie}^{7,2}$ represents the level of supervision for labeled and coarsely labeled data. 
In the ablation studies, we will investigate the effect of using different levels of supervision.
Note that we can add supervised loss on all seven levels of the taxonomy. 
However, in our experiments we did not find it useful to add losses at a level coarser than which the supervision was provided, \eg losses at the Genus level or higher when Species labels are provided.
Our method can also be extended to general hierarchical graphs such as WordNet using marginalization methods~\cite{deng2014large,samplawski2020zero}.

\subsection{Joint training with semi-supervised loss}
In addition to the hierarchical loss, we add semi-supervised losses such as consistency regularization, entropy minimization, or pseudo-labeling on the Species level for coarsely labeled data.
We select representative semi-supervised methods including Pseudo-Label, FixMatch, Self-Training with distillation, and Self-Supervised learning (MoCo) with distillation. 
We describe how we incorporate hierarchical supervisions with these methods in the following.

\paragraph{Pseudo-Label~\cite{lee2013pseudo}.}
Pseudo-label uses the model's predictions as labels if the prediction is higher than a threshold $\tau$.
Denote the pseudo-label in the Species level as $\hat{q}_i^7 = \argmax(q_i^7)$, then the loss for pseudo-label training is:
\begin{equation}
\mathcal{L} = \mathcal{L}_\text{hie}^{7,2} + \sum_{j=1}^{n}\mathds{1}\big[\max(r_i) \geq \tau\big] H (\hat{q}_i^7,q_i^7) .
\end{equation}

\paragraph{FixMatch~\cite{sohn2020fixmatch}.}
FixMatch utilized two different augmentation functions for consistency training, one with weak augmentation $\alpha(\cdot)$ and one with strong augmentation $\mathcal{A}(\cdot)$. For each coarsely labeled image $u_j$, the KL distance between the pseudo-label from weakly-augmented image $q_j = f(\alpha(u_j))$ and the prediction of strongly-augmented image $Q_j = f(\mathcal{A}(u_j))$ is minimized. 
To compute the supervised loss for labeled data, weak augmentation $p_i = f(\alpha(x_i))$ is used. 
For the supervised loss on coarsely labeled data $\mathcal{U}$, we use strongly-augmented images $Q_j = f(\mathcal{A}(u_j))$ since weakly-augmented images are only used for generating pseudo-labels without back-propagation.
The final loss is:
\begin{equation}\label{eq:fixmatch_hie}
\mathcal{L} = 
\underbrace{
\sum_{i=1}^{m} H(y_i^7, p_i^7) +
\sum_{j=1}^{n} H(y_j^2, Q_j^2)
}_{\mathcal{L}_\text{hie}^{7,2}} +
\sum_{j=1}^{n}\mathds{1}\big[\max(q_i^7) \geq \tau \big] H (\hat{q}_i^7,Q_i^7).
\end{equation}

\paragraph{Self-Training.}
We use model distillation~\cite{hinton2015distilling} as the self-training method.
Specifically, we first train a teacher model using labeled data for supervised learning $\mathcal{L} = \sum_{i=1}^{m} H(y_i^7, p_i^7)$.
We then train a student model using distillation loss, which is the KL distance between the \emph{logits} from the teacher and student models (denoted as $z^t$ and $z^s$).
The final loss is:
\begin{equation}\label{eq:distillation_hie}
\mathcal{L} = \mathcal{L}_\text{hie}^{7,2} + \sum_{i=1}^{n}H \left(\sigma\left(\frac{z_i^t}{T}\right), \sigma\left(\frac{z_i^s}{T}\right)\right),
\end{equation}
where $\sigma(\cdot)$ is the softmax function and $T$ is the temperature parameter.

\paragraph{Self-Supervised Learning (MoCo)~\cite{he2020momentum}.}
We use Momentum Contrastive (MoCo)~\cite{he2020momentum} for self-supervised learning on the union of labeled and coarsely labeled data ($\mathcal{L}\cup\mathcal{U}$). Specifically, MoCo uses contrastive learning to minimize the representations of two different augmentations of an image. 
Denote $q=f(x)$ as the representation of an image $x$ and $k^{+}$ as the positive sample, which is another augmentation of the same image $x$.
The negative samples $k_i^{-}$ are sampled from a memory bank.
The InfoNCE~\cite{oord2018representation} loss for the query $q$ is:
\begin{equation}\label{eq:moco_hie}
\mathcal{L}_q = -\log\frac{\exp\left( q \cdot k^{+} / T \right)}{\exp( q \cdot k^{+} / T) + \sum_i^K \exp(q \cdot k_i^{-} / T )},
\end{equation}
where $T$ is a temperature hyper-parameter. The encoder for the memory bank is updated based on momentum of the encoder $f(\cdot)$ to stabilize the training. 
After the self-supervised pre-training is complete, we replace the MLP layers (after global pooling) with a linear projection layer and fine-tune the entire model using our supervised hierarchical loss (Eq.~\ref{eq:hierarchy}).
Alternatives to MoCo such as SimCLR~\cite{chen2020simple} and BYOL~\cite{grill2020bootstrap} could be used, but their effect is somewhat mitigated by the fact that ImageNet pre-training is still far more effective than self-supervised learning on fine-grained domains. However, the impact may be higher when training from scratch.

\paragraph{MoCo + Self-Training.}
This method combines the previous two methods, which is similar to the setting proposed in  Chen~\etal~\cite{chen2020big}. The MoCo pre-trained model is followed by supervised fine-tuning on the labeled set to obtain the teacher model. This is then used to self-train a student model using the distillation loss in Eq.~\ref{eq:distillation_hie}. 

\section{Experiments}\label{sec:exp_hie}

\subsection{Experimental settings}

\begin{table}[t!]
\setlength{\tabcolsep}{7pt}
\renewcommand{\arraystretch}{1.1}
\centering
\begin{tabular}{c c c c c}
\toprule
\textbf{Kingdom}&	\textbf{Phylum}&	    \textbf{$C_{in}$}&	\textbf{$C_{out}$}\\
\midrule
\multirow{4}{*}{\shortstack{Animalia\\(1294)}}&   Mollusca&   11 &    24\\
&     Chordata&   113 &   228\\
&    Arthropoda& 301 &   605\\
&    Echinodermata& 4 &    8\\
\midrule
Plantae&    Tracheophyta&  336& 674\\
(1028)&     Bryophyta& 6&   12\\
\midrule
Fungi&      Basidiomycota& 29&  58\\
(117)&      Ascomycota& 10&    20\\
\bottomrule
\multicolumn{2}{c}{Total classes} & 810 & 2439\\
\bottomrule
\end{tabular}
\quad
\begin{tabular}{cc}
    \toprule
    \textbf{Taxonomy} & \textbf{\#Classes in $C_{in}$}\\
    \midrule
    Kingdom	&3\\
    Phylum	&8\\
    Class	&29\\
    Order	&123\\
    Family	&339\\
    Genus	&729\\
    Species	&810\\
    \bottomrule
\end{tabular}
\vspace{0.05in}
\caption{\textbf{Statistics of the Semi-iNat dataset~\cite{su2021semi_iNat}.}
\textbf{Left: Number of classes under each Kingdom and Phylum.} 
The species of Semi-iNat come from 3 Kingdoms and 8 Phyla. In each Phylum, one-third of the species are used for in-class species $C_{in}$ and the rest are used for out-of-class species $C_{out}$.
\textbf{Right: Number of classes in each level of the taxonomy.}
}
\label{tab:taxa}
\end{table}

\paragraph{Dataset.} We use the Semi-iNat dataset~\cite{su2021semi_iNat} from the semi-supervised challenge~\cite{semi_inat_challenge} at the FGVC8 workshop. 
The dataset contains 810 in-class species ($C_{in}$) and 1629 out-of-class species ($C_{out}$) from 3 different Kingdoms. 
The fully labeled data come from in-class species while the coarsely labeled data are drawn from in-class and out-of-class species, denoted as $U_{in}$ and $U_{out}$ respectively.
Table~\ref{tab:taxa} shows the statistics of the dataset and Figure~\ref{fig:main} left shows the taxonomy.

\paragraph{Training details.}
We use ResNet-50~\cite{he2016identity} as our backbone model and an input size of 224$\times$224 for all the experiments.
For all the methods except for MoCo and FixMatch, we use SGD with a momentum of 0.9 to train the model with 100k (from scratch) or 50k iterations (from expert models).
The batch size is 60 for training supervised baselines. For semi-supervised methods, we sample 30 images each from labeled and coarsely labeled data with a total batch size of 60.
The learning rate is searched within $[0.001, 0.03]$, and the weight decay is set for either $0.001$ or $0.0001$. The hyper-parameters are set using the validation set of Semi-iNat. This set is not included for the supervised loss but is used for training MoCo. 

For MoCo, we follow the setting of MoCov2~\cite{chen2020improved} and use a batch size of 2048 negative samples.
The training is done with a learning rate of 0.03 and 0.0003, and for 800 and 200 epochs, for training from scratch and from expert models respectively.

For FixMatch, we follow the original setting and use RandAugment~\cite{cubuk2020randaugment} for augmentation. 
Due to the hardware constraints, we were limited to a batch size of 32 for labeled data and 160 for coarsely labeled data for training with 4 GPUs. When training from scratch we use a learning rate of 0.03 for 200k iterations; when training from expert models we use a learning rate of 0.001 for 100k iterations. 
The threshold for FixMatch is set as $0.8$ for all our experiments.

\definecolor{mygray}{gray}{0.7}
\newcommand\grey[1]{\textcolor{mygray}{#1}}
\newcommand\best[1]{\textbf{\textcolor{teal}{#1}}}
\newcommand\up[0]{\textbf{\textcolor{green}{$\uparrow$}}}
\newcommand\down[0]{\textbf{\textcolor{magenta}{$\downarrow$}}}
\begin{table}[t!]
  \setlength{\tabcolsep}{10pt}
  \renewcommand{\arraystretch}{1.2}
  \centering    
  \begin{tabular}{c c | c c | c c }
    \toprule
    & \textbf{Method} & \multicolumn{2}{c|}{\textbf{from scratch}} & \multicolumn{2}{c}{\textbf{from ImageNet}} \\
    & ( Phylum Supervision  $\rightarrow$ ) & w/o & w/ & w/o & w/ \\
    \midrule
    \multirow{6}{*}{\rotatebox[origin=c]{90}{${U}_{in}$}} & Supervised Baseline & 18.5 & 21.7\up & 40.4 & 46.6\up  \\
    & Pseudo-Label~\cite{lee2013pseudo} & 18.6 & 22.7\up & 40.3 & 44.9\up\\
    & {FixMatch}~\cite{sohn2020fixmatch} & 15.5 & 25.7\up & \best{44.1} & \best{47.9}\up  \\
    & Self-Training & 20.3 & 23.7\up & 42.4 & 44.8\up \\
    & MoCo~\cite{he2020momentum} & 30.2 & 33.5\up & 41.7 & 41.9\up\\
    & MoCo + Self-Training~\cite{su2021realistic} & \best{32.0} & \best{35.4}\up & 42.6 & 45.8\up\\
    \bottomrule
  \end{tabular}
  \caption{\textbf{Results of adding Phylum supervision on the Semi-iNat dataset.} Adding hierarchical loss at the \emph{Phylum} level 
  improves the Baseline and all the Semi-Supervised methods when images come from $U_{in}$, \ie, images with species labels corresponding to those in the test set $C_{in}$.The best methods are shown in \best{teal}. } 
  \label{tab:benchmark_hie_in}
\end{table}

\subsection{Using Phylum level supervision}
We first consider the setting where the coarsely labeled images are within the set of labeled species ($U_{in}$). In \S~\ref{sec:out} we will analyze the effect and utility of adding images of novel species from the same coarse categories.
Models are initialized randomly or from an ImageNet pre-trained model. 
For each setting, the Baseline supervised learning and five Semi-Supervised learning methods are evaluated. We then analyze the effect of adding hierarchical loss.

\begin{figure}[t!]
\centering
\includegraphics[trim={1.2cm 0.45cm 1cm 0},clip,width=0.49\linewidth]{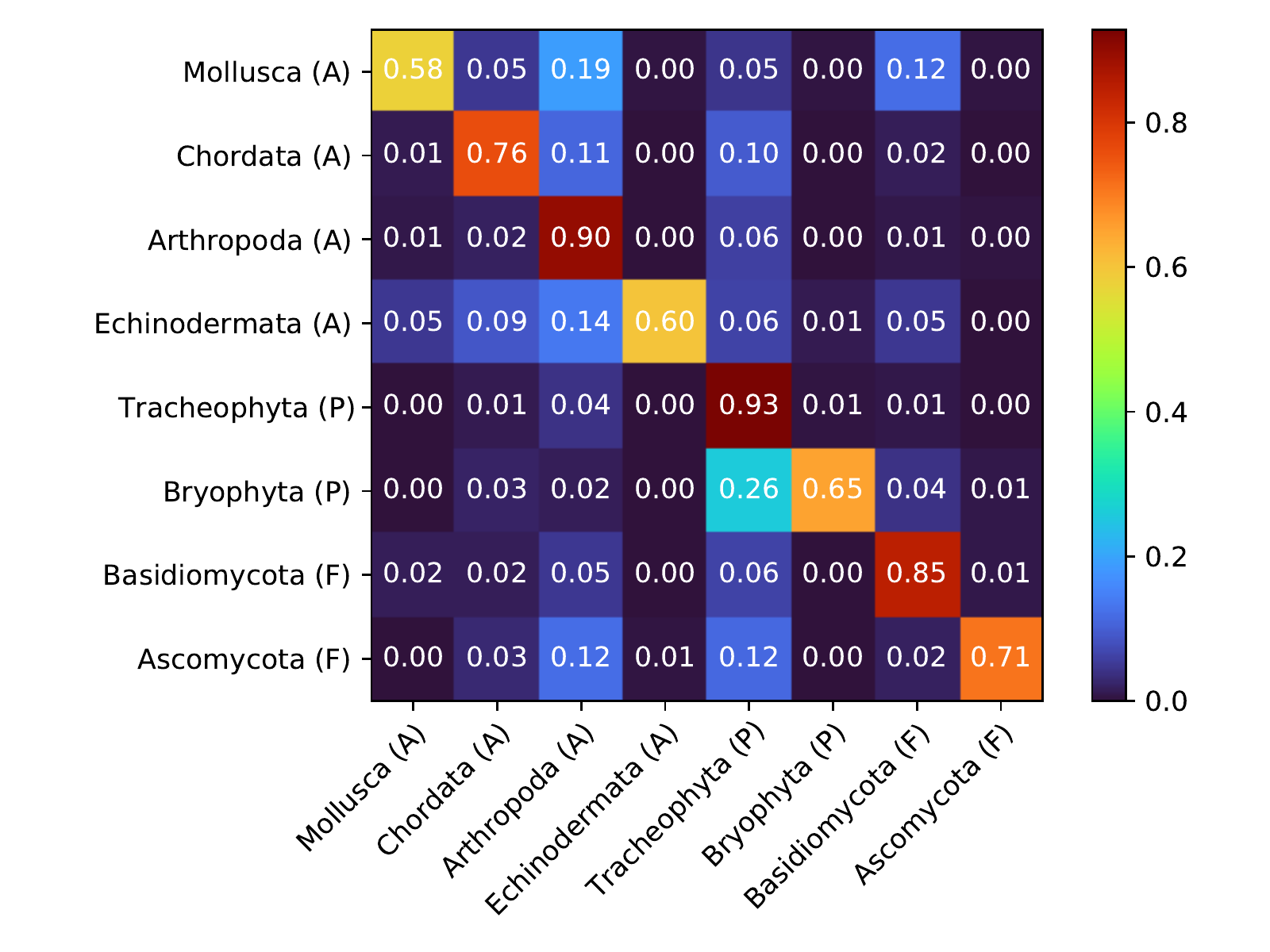}
\includegraphics[trim={1.2cm 0.45cm 1cm 0},clip,width=0.49\linewidth]{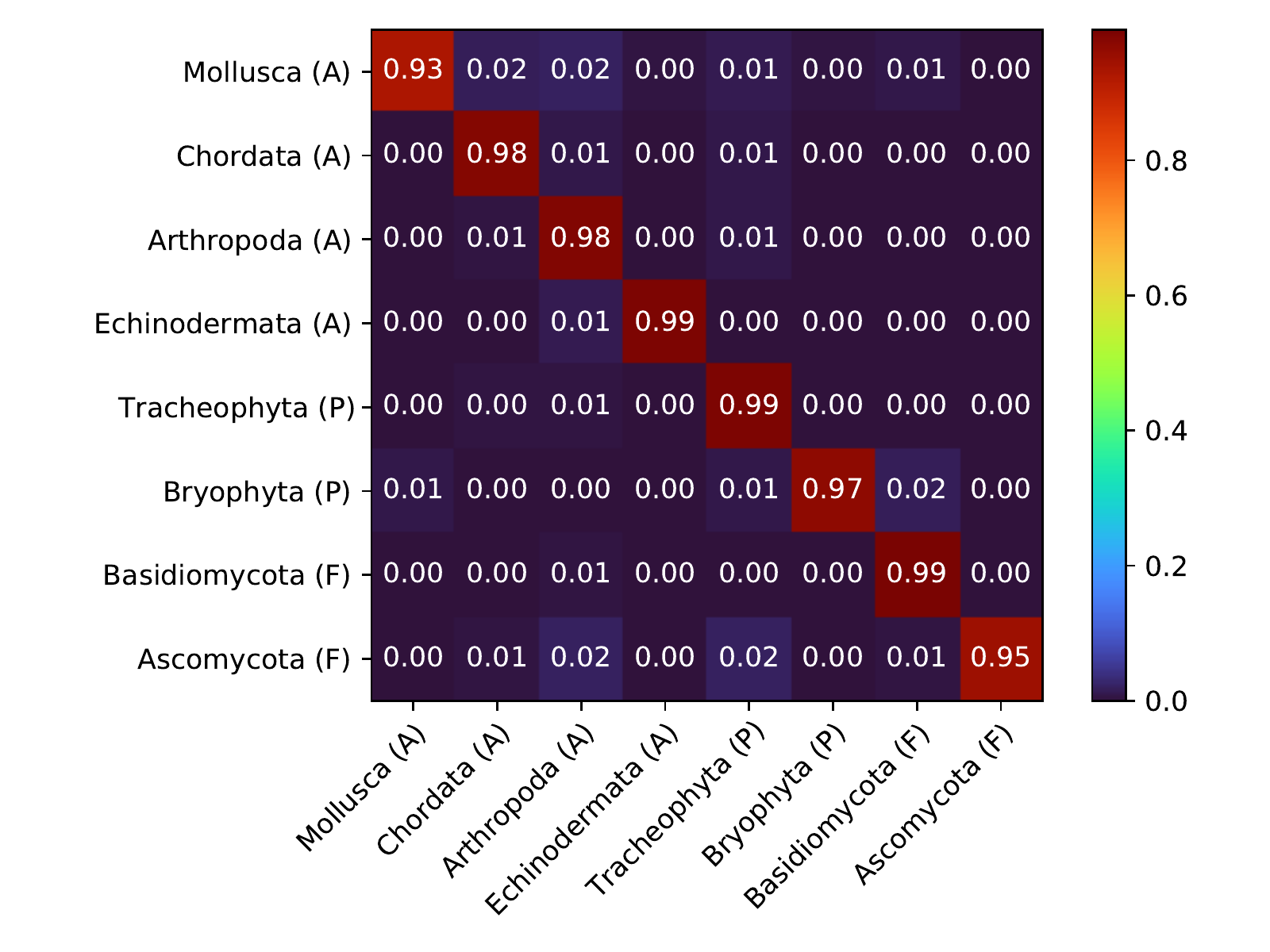}
\caption{\textbf{Confusion matrices on the Phylum level.} \textbf{Left:} Supervised Baseline model \emph{without} coarse-label supervision. \textbf{Right:} FixMatch + hierarchical loss on the Phylum level. Combining semi-supervised methods and hierarchical supervision on coarsely labeled data reduces the confusion between Phyla.}
\label{fig:confusion_hie}
\end{figure}

Results are presented in Table~\ref{tab:benchmark_hie_in}.  Adding a hierarchical loss gives almost a 10\% improvement for FixMatch and 3\% for all other methods when models are trained from scratch. 
When initialized with ImageNet pre-trained models, adding hierarchical loss provides  2-6\% improvements in top-1 accuracy except for the self-supervised training method (MoCo). 
Figure~\ref{fig:confusion_hie} shows the confusion matrices at the Phylum level for models trained with and without hierarchical supervision.
Hierarchical supervision sensibly reduces confusion among the four Phyla within the animal (A) Kingdom (\eg, Antropoda vs.~Echinodermata), as well as at the Plant (P) Kingdom. The combined effect of hierarchical supervision and semi-supervised learning represents an overall improvement from 40.4\% to 47.9\%.

\subsection{Using different levels of supervision}
In this section, we consider supervision across different levels of the taxonomy on top of FixMatch and the supervised baseline.
For example, if we have the labels in the Order level for coarsely labeled data, \ie $(u_j,y_j^4) \in \mathcal{U}$, then the hierarchical loss becomes $\mathcal{L}_\text{hie}^{7,4}$.
We use the ground-truth labels (on the Species level) of the coarsely labeled data which are released after the competition ends~\cite{semi_inat_challenge} for the analysis.
As shown in Figure~\ref{fig:main} right, we can see that finer-grained labels improve performance, but this also requires more annotation effort.
The number of classes at each level in the taxonomy shown in Table~\ref{tab:taxa} provides a proxy for the annotation effort.
Even incorporating labels on the Kingdom level (which only has three categories) leads to improvements, 
while Class level (29 categories) improves the performance of FixMatch from 44.1\% (FixMatch + None) to 51.8\% (FixMatch + Class). 
In these experiments, the coarsely labeled set $\mathcal{U}_{in}$ on which the semi-supervised losses are constructed in Eq.~\ref{eq:fixmatch_hie} are kept the same.
Interestingly, even when the method is fully supervised, these self-supervised losses are useful, and provide an improvement over the fully-supervised Baseline. 
This is perhaps not as surprising as previous works have noted that the self-supervised losses improve few-shot learning~\cite{zhai2019s,Su2020When,gidaris2019boosting}.

\begin{table}[t!]
  \setlength{\tabcolsep}{10pt}
  \renewcommand{\arraystretch}{1.2}
  \centering    
  \begin{tabular}{c c | c c | c c }
    \toprule
    & \textbf{Method} & \multicolumn{2}{c|}{\textbf{from scratch}} & \multicolumn{2}{c}{\textbf{from ImageNet}} \\
    & ( Phylum Supervision $\rightarrow$ ) & w/o & w/ & w/o & w/ \\
    \midrule
    \multirow{6}{*}{\rotatebox[origin=c]{90}{${U}_{in}+{U}_{out}$}} & Supervised Baseline & 18.5 & 20.5\up & 40.4 & \best{45.6}\up \\
    & Pseudo-Label~\cite{lee2013pseudo} & 18.8 & 21.2\up & 40.3 & 44.0\up\\
    & {FixMatch}~\cite{sohn2020fixmatch} & 11.0 & 21.1\up & 38.5 & 41.1\up \\
    & Self-Training & 19.7 & 23.3\up & \best{41.5} & 44.1\up \\
    & MoCo~\cite{he2020momentum} & 31.8 & 29.4\down & 40.8 & 39.3\down \\
    & MoCo + Self-Training~\cite{su2021realistic} & \best{32.9} & \best{35.4}\up & \best{41.5} & 42.6\up \\
    \bottomrule
  \end{tabular}
  \caption{\textbf{Results on the Semi-iNat dataset with novel classes.} 
  The presence of novel classes $C_{out}$ when adding $U_{out}$ reduces the performance of most methods compared to using $U_{in}$ shown in Table~\ref{tab:benchmark_hie_in}. However, adding hierarchical loss provides improvements, both when training from scratch or from ImageNet. We also find that MoCo+Self-Training is the most robust when training from scratch, and no method is able to improve over the supervised Baseline when ImageNet pre-training is used. The best methods are shown in \best{teal}. 
  } 
  \label{tab:benchmark_hie_out}
\end{table}

\subsection{Effect of domain shift}\label{sec:out}
Next, we consider the case when there is a domain shift, \ie training with data from $U_{in}+U_{out}$. 
The results are shown in Table~\ref{tab:benchmark_hie_out}.
When there is no hierarchical loss, the performances of semi-supervised methods all drop except for MoCo. 
Adding hierarchical loss improves all the methods except for MoCo, though the improvement is less compared to Table~\ref{tab:benchmark_hie_in}. 
Surprisingly, when training from the ImageNet model, no semi-supervised method improves over the supervised Baseline trained with the hierarchical loss.
In particular, FixMatch is less robust to the domain shift, echoing the findings in~\cite{su2021realistic}. 
One potential reason is the large domain shift between $U_{in}$ and $U_{out}$ the dataset: although the out-of-domain classes are drawn from the same Class in the taxonomy, the appearances can be significantly different. 
To alleviate the effect of domain shift, we propose to filter the out-of-domain data by measuring the uncertainty in the model's predictions. 
This step is performed before any semi-supervised training.
Similar to pseudo-labeling, we first use the baseline supervised model to generate predictions for images in $U_{in}+U_{out}$, then check if the maximum probability is greater than $\tau=0.8$.
Note that the set of out-of-domain classes ($C_{out}$) remains unknown.
We further check if the predicted labels match the provided coarse labels at the Phylum level to filter out-of-domain images.

After filtering out-of-domain data, we train FixMatch using the selected coarse-labeled images.
Using the supervision at the Phylum level obtains 42.0\% accuracy compared to 41.1\% without any domain selection. 
This allows us to use uncurated data, though the performance is lower than having only in-domain data (47.9\%). 
Our proposed approach is simple, and there is significant room to improve this using better out-of-domain detection techniques and modeling the taxonomy. Nevertheless, the problem is challenging in fine-grained domains.

\section{Conclusion}
We showed that coarse labels improve semi-supervised learning on fine-grained image classification tasks. Thus, collecting coarse labels might provide a practical way to train models on new domains where a vast number of images are readily available, but labeling effort is expensive.
We also found that out-of-domain (or novel class) data leads to a drop in performance. Hierarchical labels help, but the task of selecting relevant images can be challenging in fine-grained domains. 
Our model based on uncertainty and filtering using coarse labels only provides modest improvements.
Improved techniques for detecting out-of-domain data combined with taxonomy-aware user input could provide further benefits.

\paragraph{Acknowledgements.} This project is supported in part by NSF \#1749833 and was performed using high performance computing equipment obtained under a grant from the Collaborative R\&D Fund managed by the Massachusetts Technology Collaborative.

\bibliography{egbib}
\end{document}